\documentclass[conference]{IEEEtran}
\usepackage{cite}

\ifCLASSINFOpdf
\else
\fi
\usepackage[utf8]{inputenc} 
\usepackage[fleqn]{amsmath}
\begin{document}
%
\title{A Basic Recurrent Neural Network Model}

\author{\IEEEauthorblockN{Fathi M. Salem}
\IEEEauthorblockA{Circuits, Systems, and Neural Networks (CSANN) Lab\\Department of Electrical and Computer Engineering\\
Michigan State University\\
East Lansing, Micigan 48824--1226\\
Email:salemf@msu.edu}}



%


\maketitle

\begin{abstract}
We present a model of a basic recurrent neural network (or bRNN) that includes a separate linear term with a slightly "stable" fixed matrix to guarantee bounded solutions and fast dynamic response. We formulate a state space viewpoint and adapt the constrained optimization Lagrange Multiplier (CLM) technique and the vector Calculus of Variations (CoV) to derive the (stochastic) gradient descent. In this process, one avoids the commonly used re-application of the circular chain-rule and identifies the error backpropagation with the co-state backward dynamic equations. 
We assert that this bRNN can successfully perform regression tracking of time-series. 
Moreover, the ``vanishing and exploding'' gradients are explicitly quantified and explained through the co-state dynamics and the update laws.
The adapted CoV framework, in addition, can correctly and principally integrate new loss functions in the network on any variable and for varied goals, e.g., for supervised learning on the outputs and unsupervised learning on the internal (hidden) states.
\end{abstract}


%


\section{Introduction}

The so-called simple Recurrent Neural Networks (sRNN) have been reported to be difficult to train by the stochastic gradient descent, see \cite {IEEEhowto:lecun_deep_2015}, \cite {IEEEhowto:Chung_2014} and the references therein. This has spurred developments of new training approaches, modifications, and more complex architectures to enable convergent training and learning in state-of-the-art applications 
 \cite {IEEEhowto:Chung_2014}, \cite{IEEEhowto:Joze_2015}, \cite {IEEEhowto:Zhou_minimal_2016}. Recent trends have focused on reducing the computational load and complexity while preserving their demonstrated learning capabilities of sequences and time-series  \cite{IEEEhowto:Joze_2015} 
and \cite {IEEEhowto:Le_2015}.
\\
In this work, we define a basic recurrent neural network (bRNN) as a model that can achieve regression (i.e., output tracking) of real (i.e. analog) output values. Using the classical Lagrange multiplier approach for non-convex constrained optimization, we obtain a bounded input-bounded-output (BIBO) stable flexible recurrent neural network that can be trained to learn (discrete) categories or (continuous) regression profiles. For ease of presentation, we adopt a tutorial style and relegate most derivations to the appendices. 
\\
This bRNN has the following unique attributes: 
\\
\begin{itemize}  
\item Stable behavior without a need of additional ``gating" networks. Gating networks would at least double the number of the overall adaptive parameters in the overall RNN. In addition, no need for special caps on the growth of the gradient signals. 
\\
\item Specify a predictive state space formulation. Specifically, the network takes input data and state values at one index (or time) value, say $k$ and produces a state value and/or output at the next index, say $k+1$.
\\
\item Update law derivations, including the backpropagation network, are shown to follow easily using the Lagrange Multiplier Constrained Optimization method. One identifies the forward and backward propagation networks and the split boundary conditions in a principled way. The split boundary conditions are the initial state and final co-state, and they are the fundamental reason behind the state-forward and costate-backward propagating dynamic processing for the stochastic gradient descent. 	
\\
\item In this framework, extensions and definitions of the output or state loss functions can easily be incorporated without complex re-derivations. E.g., each loss function can be independently specified to be a supervised cost (with a given reference/target), or an unsupervised loss function, e.g., minimizing the entropy or maximizing sparsity in an internal representation (state) vector. 

\end{itemize}

\section{simple Recurrent Neural Networks}

The simple recurrent neural network (sRNN) model is expressed as: 
\begin {align} 
h_{k} &= \sigma_{k}  ( U  h_{k-1} + W  s_{k} +  b) , && ~~ k=0,..., N ~ \label {eq: eqn1} 
\end {align}
where  $k$ is the discrete (time) index, and $N$ is the final finite-horizon time,  $s_{k}$ is the m-d input vector sequence, and $h_{k}$ is the n-d output via the nonlinear function $\sigma_{k}$. Here,  
$\sigma_{k}$ is a general nonlinear function which may be specified to be the logistic function $sigm$ or the hyperbolic tangent $tanh$ as is common in the literature (see \cite{IEEEhowto:lecun_deep_2015}- \cite  {IEEEhowto:Zhou_minimal_2016}, and the references therein), or even as the rectified linear unit $reLU$ as in \cite {IEEEhowto:Le_2015}. The non-indexed parameters, to be determined by training, are the $n \times n$-matrix $U$, the $n \times m$-matrix $W$, and the $n \times 1$-vector bias $b$. This model is a discrete nonlinear dynamic system, see, e.g., \cite {IEEEhowto:Chung_2014}.

Using variants of stochastic gradient descent to train the sRNN for long-term dependencies or sequences, it has been widely reported that the sRNN may exhibit the so-called ``vanishing gradient" or the ``exploding gradient," see \cite{{IEEEhowto:Chung_2014}, {IEEEhowto:Le_2015}} and the references therein.

The Long Short-Term Memory (LSTM) network was developed to overcome difficulties in training sRNNs. They have demonstrated their state-of-the-art performance in many applications, e.g., in mapping input sequence to an output sequence as in speech recognition, machine translations, and language modeling, see [1-3]. In a nutshell, the three gating control signals in the LSTM network, each roughly replicates a sRNN and bring a four-fold increase in parameters in comparison to parameters of the same  state-size sRNN. More recently, there has been interest in (i) determining the optimal RNNs with decreased computational load and parameters. e.g., the GRU architecture \cite {IEEEhowto:Chung_2014}, reduces the gating signals to two and thus parameters become three folds in comparison to the parameters in the same state-size sRNN. Moreover, the recent MGU architecture \cite {IEEEhowto:Zhou_minimal_2016} further reduces the number of independent control gates to one and thus the parameters become two-folds in comparison to the parameters in the same state-size  sRNN. There is also interest in reviving simple RNNs with modifications in nonlinearity-type, and an approach to careful training which has introduced  the IRRN design, i.e., sRNN with rectified Linear Units (ReLUs) for the nonlinearity, an initialization of the $U$ matrix in eqn (\ref {eq: eqn1}) at the identity, and setting a cap limit on the resulting gradient computation \cite {IEEEhowto:Le_2015}.
\\
\\
\textbf{\underline{Remark II.1:}} From a system and signal processing viewpoints, the model in eqn 
(\ref {eq: eqn1}), without the nonlinearity ($\sigma_k$), represents a filtering that can be expressed in the (frequency) Z-transform as follows: 
\begin {align} 
H(z) &=\left[I-Uz^{-1} \right]^{-1} \left(  W S(z) ~+~ b\right ) ~  \nonumber \\
&= z \left[z I-U \right]^{-1} \left(  W S(z) ~+~ b\right )  \nonumber 
\end {align}
In general, this represents a high-pass filtering, and thus all high frequency input components, including noise, will pass through to the output. This motivates one to revise the indexing at the input and to include an  output layer as part of the model in order to realize a flexible filtering, and specifically to be able to realize a bandpass filtering with stable poles after training. This description of course is valid only for linear systems, and it serves only as a motivation of the design of the nonlinear system. 
\\
\\
To motivate the proposed model in the next section, we can now re-write this sRNN model as:
\begin {align} 
x_{k} &= U  h_{k-1} + W  s_{k} +  b , && ~~ k=0,..., N ~ \label {eq: eqn01} \\
h_{k} &= \sigma_{k}  (x_{k} ), &&~~ k=0,..., N  \label {eq: eqn02} 
\end {align}
\\
where we explicitly identified the state vector $ x_k$.  Eqn (\ref {eq: eqn01}) captures the stability behavior. The (local) stability properties of the network is determined by the matrix $U h_{k-1}^{'}$, where  $h_{k-1}^{'}$ is the diagonal matrix derivative of the vector  $h_{k-1}$. However, the matrix $U$ is typically initialized randomly small and is adaptively changing over the training iterations. That is, $U h_{k-1}^{'}$ may become unstable and may lead to unstable behavior of the dynamic network during the training phase! If $Uh_{k-1}^{'}$ is an unstable matrix (meaning that at least one of its eigenvalues has a modulus greater than unity), the nonlinear dynamic system \textit{may} become unstable which may lead to $x_k$ growing without bound. Even though $h_k$ is bounded when using the compressive nonlinearity $\sigma_k$, the state $x_k$ can become unbounded and thus the dynamic system may in fact become unstable in the sense that $x_k$ grows unbounded. Of course, Liapunov theory and Liapunov functions constitue one approach to definitively confirm the stability of such a nonlinear system. 
\\
\\
It would be beneficial to add a (stable) linear term in the state eqn (\ref {eq: eqn01}) that bounds the growth of $x_k$ to within a bounded region. In Appendix I, we include the analysis based on Liapunov theory to show that such dynamic system would indeed be bounded to a region in the ($x_k$-)  state space. In addition, one includes a linear output layer in the modeling to allow for flexible filtering which is intended to render the recurrent network with its output layer a  general bandpass filter. Such a filter would have the equivalent of poles and zeros that produce, after training,  a linearizable system with an equivalent \textit{proper} stable transfer function. These notions are incorporated into the basic RNN model defined next.  
\\

\section{Basic Recurrent Neural Networks: bRNN-- the general case}
In contrast to the sRNN, see eqns (\ref {eq: eqn01})-(\ref {eq: eqn02}), now consider this extended recurrent neural network model: 
\begin {align} 
x_{k+1} &= A x_{k} + U  h_{k} + W  s_{k} +  b , &&~~ k=0,..., N-1 ~ \label {eq: eqn11} \\
 h_{k} &= \sigma_{k}  (x_{k} ), &&~~ k=0,..., N  \label {eq: eqn12} \\
 y_{k}  &= V h_{k} + D  s_{k} + c , &&~~ k=0,..., N  \label {eq: eqn13}
\end {align}
where, as before, $k=0,...,N$, is the discrete (time) index, $s_{k}$ is the m-d input vector sequence, $x_{k}$ is the n-d state vector, $h_{k}$ is the n-d nonlinear function of the state vector which is labeled as the hidden unit, and $y_{k}$ is the r-d corresponding output vector sequence. For this neural system, 
eqn (\ref{eq: eqn11}) includes the dynamic transition from the index step $k$ (all variables on the right-hand side have the same index $k$) to the state at the next step $k+1$. Eqn (\ref{eq: eqn12}) represents the static nonlinear state transformation which may include any of the common nonlinearity, e.g., the logistic function, hyperbolic tangent, reLU, etc.. Eqn (\ref{eq: eqn13}) represents the static output equation. 

This neural model extends the simple recurrent neural networks (sRNN)  by adding \textit{a linear state} term to the dynamic equation, eqn ((\ref{eq: eqn11}), via the matrix $A$, and also by adding \textit{ the direct input} term in the output equation, eqn (\ref{eq: eqn13}), via the matrix $D$. The dimensions of the parameters are obvious to achieve compatibility of the equations. Specifically, $A$ and $U$ are $n \times n$, $W$ is $n \times m$, $b$ is $n \times 1$, $V$ is $r \times n$, $D$ is $r \times m$, and $c$ is $r \times 1$.

In eqn (\ref{eq: eqn11}), the state matrix $A$ is set to be constant with eigenvalues having moduli within (or on) the unit circle. As a special case, one may choose the eigenvalues to be distinct, real, or random, with amplitude $\leq 1$. For large scale models, it may be computationally easier to choose $A=\alpha I,0< ~\alpha  \leq 1 $. The state matrix $A$ must  be a \textit{stable} matrix. In eqn (\ref{eq: eqn13}), the direct  input term enriches the model since the state equation can only produce transformations of delayed versions of the input signal but not an  instantaneous input.

The parameters in the dynamic equation, eqn  (\ref{eq: eqn11}), enumerated in the matrices $U, W$, and the bias vector $b$, can be represented by a single vector parameter $\theta$, whereas the parameters of the output equation eqn (\ref{eq: eqn13}), enumerated in the matrices $V, D$, and the bias vector $c$, can be represented by the single vector parameter $\nu$. We shall also index these parameters by the index $k$ in order to facilitate the upcoming analysis. Thus we shall represent the dynamic equations as follows: 
\begin {equation}
\begin {aligned}
x_{k+1} &=f^{k}(x_{k} , h_{k}  , s_{k} ,  \theta_{k} ) \nonumber \\
&= A x_{k} + U_{k} h_{k} + W_{k}  s_{k} +  b_{k} , &&~ k=0,..., N-1 ~~\\
 h_{k} &= \sigma_{k}  (x_{k} ), ~&&~ k=0,..., N \\
 y_{k}  &=  g^{k}(h_{k}  , s_{k} ,  \nu_{k} ) \nonumber \\
&=V_{k} h_{k} + D_{k}  s_{k} + c_{k} , ~&&~ k=0,..., N \label {eq: eqn203}
\end {aligned}
\end {equation}
where $s_{k} , k=0,...,N$ is the sequence of input signal, and $k$ is the time index of the dynamic system. In the case of supervised adaptive learning, we associate with each sample $s_{k}$ the desired sequence label $d_{k}$.
\\
\\
The cost or loss function is thus given in general as a function on the final time output $y_{N}$ as well as the variables at the  intermediate times within $o \leq k <  N$:
\begin {equation}
\begin {aligned}
J_{o} (\theta_{o}, ..., \theta_{N-1} , \nu_{o} ,..., \nu_{N}) &= \phi (y_{N}) +  L^N (\nu_{N}) \nonumber  \\
&~~\sum_{k=o}^{N-1}  L^{k} (y_{k}, x_{k} , h_{k} ,  \theta_{k} , \nu_{k}) 
\end {aligned}
\end {equation}
where the general loss function includes all variables and parameters.  

Following the calculus of variations in constrained optimization and the Lagrange multiplier technique, see 
e.g., \cite {IEEEhowto:Bryson_1975} and \cite {IEEEhowto:Lewis_2012}, one can define the Hamiltonian at each time step $k$ as
\begin {equation} 
\begin {aligned}
H^{k}= L^{k} (y_{k}, x_{k} , h_{k},  \theta_{k} , \nu_{k} )+ & ( \lambda_{k+1})^{T} f^{k} (x_{k} , h_{k} , s_{k}, \theta_{k} ),  \nonumber  \\ 
&~~~ ~~~~~ i=1,..., N-1
\end {aligned}
\end {equation} 
\\
where the sequence $ \lambda_{k}, 0 \leq k \leq N$ of the Lagrange multipliers would become the co-state with dynamics generated as in constrained optimization and optimal control, 
see \cite {IEEEhowto:Bryson_1975} and \cite {IEEEhowto:Lewis_2012}. 
The state equations and the co-state equations are reproduced from optimization, similar to  optimal estimation and control procedures with minor adjustments.  We apply the (stochastic) gradient descent to all the parameters.  Recall that the parameters $\theta_{k}$ represent the elements of the matrices $U_{k} , W_{k}$ and bias vector $b_{k}$. Similarly, the parameters $\nu_{k}$ represent the elements of the matrices $V_{k} ,  D_{k}$ and the bias vector $c_{k}$.

Thus, the state equations representing the network is reproduced as: 
\begin {align}
 x_{k+1} &=
\frac{\partial  H^{k}}{\partial \lambda_{k+1}}   = f^{k} \nonumber \\
&=  A x_{k} + U_{k} h_{k} +  W_{k} s_{k}  + b_{k} ,    \label {eq: eqn81}
&~ i=0,..., N-1  \nonumber  \\  
\end {align}
where the output layer is 
\begin {align}
 y_{k}  =  V_{k} h_{k} + D_{k}  s_{k} + c_{k} , ~&&~ k=0,..., N \label {eq: eqn203}
\end {align}
\\
The co-state dynamics are generated as: 
\begin {align}
\lambda_{k}  &= \frac{\partial  H^{k}}{\partial x_{k}}  \nonumber \\
&= \frac{\partial  L^{k}}{\partial x_{k}}  ~+~ 
\left(\frac{\partial f^{k} }{\partial x_{k}} \right)^{T}  \lambda_{k+1}, ~~\label {eq: eqn303} ~ 0 < k  \leq N-1
\end {align}
The co-state dynamics are linearized (sensitivity) equations along the state trajectory. 
\\
\\
The gradient change in the parameters (within the state equation) at each time step $k$ are: 
\begin {align}
\Delta \theta_{k} &= - \eta
\frac{\partial  H^{k}}{\partial \theta_{k}}   \nonumber \\
&= - \eta
\left( \frac{\partial  L^{k}}{\partial \theta_{k}}  ~+~ 
\left(\frac{\partial f^{k} }{\partial \theta_{k}} \right)^{T}  \lambda^{k+1} \right)  \label {eq: eqn101}
\end {align}
where $\eta$ is a general (sufficiently small) learning rate. Similarly, the gradient change in the parameters (within the  output equation) at the time step $k$ are: 
\begin {align}
\Delta \nu_{k} = - \eta
\frac{\partial  H^{k}}{\partial \nu_{k}}   = - \eta
\left( \frac{\partial  L^{k}}{\partial \nu_{k}}   \right)  \label {eq: eqn102}
\end {align}
Note that  eqns (\ref {eq: eqn101})-(\ref {eq: eqn102})  are written as deterministic expressions for clarity; however, they should be viewed with an expectation operator applied to the righ-hand side. These changes are expressed at each time step $k$, thus there is a ``sequence" of gradient changes over the whole time horizon duration $0$ to $N$. The goal is to use these time-step changes to compute the changes to the parameters over the whole time horizon of the sequence trajectory, known as an epoch. We note that the parameters of the network are ``shared" and must remain  constant during the trajectory/epoch and can only be incremented from an epoch to the next epoch till they converge to a constant set of parameters. 
\\
\\
Finally, the split boundary conditions are: 

\begin {align}
\left( \frac{\partial  \phi}{\partial x_{N}}  - \lambda_{N} \right )^{T}  dx_{N}=0 ~\label {eq: eqn103}
\end {align}

\begin {align}
\left( \lambda_{o} \right )^{T}  dx_{o}=0 ~\label {eq: eqn104}
\end {align}
\\
It is usually assumed that the initial state stage $x_{o}$  is given or specified as constant, and thus $dx_{o}=0$. Therefore eqn (\ref {eq: eqn104}) is trivially satisfied. Also, for the boundary condition eqn
(\ref {eq: eqn103}) we take the quantity in parenthesis to equal zero. 
The boundary conditions thus become: 
\\
\\
(i) Initial boundary condition: 
\begin {align}
x_{o} ~~~
\end {align}   \\
(ii) Final boundary condition: 
\begin {align}
\lambda_{N} = \frac{\partial  \phi}{\partial x_{N}}   ~\label {eq: eqn105}
\end {align}
\\
Thus, given a set of parameters values, the iteration processing is to use the initial state $x_{o}$, the input signal sequence $s_{k}$ and the corresponding desired sequence $d_{k}$ to generate the  state and output sequences (forward in time) up to the final time$N$. Then, compute the final boundary of the co-state using eqn (\ref {eq: eqn105}) and subsequently generate the co-state sequence $\lambda_k$ of eqn (\ref {eq: eqn303}) backward in-time. For each iteration, or a batch of iterations, all parameters need to be updated, and continue doing so according to some stopping criterion. In principle, this is the core parameters iteration procedure in training neural networks. After each iteration is completed, and/or after the stopping criterion is met, one may use the achieved fixed parameters in the recurrent network, eqns (\ref {eq: eqn81})-(\ref {eq: eqn203}), for evaluation, testing, or processing.  
\\

\section{Basic Recurrent Neural Networks: bRNN--special case}

We now provide specific details of the bRNN model and its parameter updates.
We basically specify the components of the loss/cost function and the bounday conditions, namely the initial condition for the state forward dynmaics and final condition for the co-state backward dynamics. 
\\
\\
The final-time component of the loss function $ \phi (y_{N})$ can be specified for supervised learning as, e.g., an $L_{2}$ norm of an error quantity at the final time $N$. Specifically, 
\begin {align}
\phi (y_{N}) &= \frac{1}{2} ||y_{N} -d_{N}||^{2}_{2} \nonumber  \\
&= \frac{1}{2} (y_{N} -d_{N})^T (y_{N} -d_{N})  \label {eq: eqn403}
\end {align}
\\
And thus its derivative is calculated from vector-calculus to produce the final condition of the co-state as: 
\begin {align}
\lambda_{N} = \frac{\partial  \phi}{\partial x_{N}}  ~=~  (\sigma_{N}^{'})^T    V_{N}^{T} ~ (y_{N} - d_{N})
\label {eq: eqn4403}
\end {align}
\\
where one uses the output eqn (\ref {eq: eqn203}) for the definition of $y_N$. The matrix 
$ \sigma_{N}^{'}$ is the derivative of the nonlinearity vector $\sigma_{N}$ expressed as a diagonal matrix of element-wise derivatives.  This provides the final co-state at the final time $N$.
\\
\\
\textbf{\underline{Remark IV.1:}}
Note that for computational expediency, the diagonal matrix $\sigma_{N}^{'}$ can be expressed as a vector of derivatives multiplied point-wise (i.e., a Hadamard multiplication) to the vector $y_N - d_N$. This is more commonly adopted in computational/coding implementations.
\\

The general loss function may be simplified to be a sum of separate loss functions in each vector variable, e.g., as 
\begin {align}
L^{k} (y_{k}, x_{k} , h_{k} ,  \theta_{k} , \nu_{k} ) ~&=~ L^{k} (y_{k})  \nonumber  \\
&~+ ~\beta L^{k} ( x_{k}) ~+~ \beta_{0} L^{k} ( h_{k}) \nonumber  \\
&~ +~\gamma_1  L^{k} (  \theta_{k}) ~+~ \gamma_2  L^{k} (  \nu_{k})
\end {align}
\\
where the first term is a loss function on the output at index k and is chosen to be a supervised loss function with corresponding reference or target $d_k$. One may use the scaled  $L_2$ norm of the error to be consistent with the final-time loss function in eqn (\ref{eq: eqn403}), specifically, 
\begin {align}
 L^{k} (y_{k})  &=~  \frac{1}{2} ||y_{k} -d_{k}||^{2}_{2} \nonumber  \\
&=~ \frac{1}{2} (y_{k} -d_{k})^T (y_{k} -d_{k})
\end {align}
All other terms have the tuning (or hyper-) parameters $\beta, \beta_o, \gamma_1$ and $\gamma_2$ as scaling penalty factors between $0$ and $1$ to emphasize the importance one places on these individual costs. The second two loss terms (usually, one needs to use only one or the other), are on the internal state or its hidden function. One may use either one with an unsupervised loss function,  e.g., to optimize the entropy, cross entropy, or sparsity of this internal (state) representation.  Here, we set $\beta_0$ to zero and choose the loss function to be the entropy defined as
\begin {align}
 L^{k} (x_{k})  &=~  -\ln | p_{x_k} (x_k) | \nonumber  \\
& \approx~ || x_k ||_1 ,  %
\end {align}
where one may use the $L_1$ norm to approximate the entropy (i.e., one imposes the sup-Gaussian Laplacian density function on the state vector $x_k$).  Another candidate is to use the general approximating function for sup-Gaussian densities, namely, $tanh(\alpha * x_k), ~ 1 < \alpha \leq 3$,  for further details see, e.g. \cite {IEEEhowto:Khuram_2003} and \cite {IEEEhowto:Albatain_2016}. 
\\
Finally, the loss function terms on the parameters $\theta_k$ and $\nu_k$ may be used for regularization. In that case, one provides a scaled quadratic expression for every scalar parameter. An example is to use 
\\
\begin {equation}
\begin {aligned}
 L^{k} (  \theta_{k})  &=~  \frac{1}{2} ||\theta_{k} ||^{2}_{2}    ~~0 \leq k \leq N-1 \nonumber  
\end {aligned}
\end {equation}
and for the parameters in the output layer, 
\begin {equation}
\begin {aligned}
 L^{k} (  \nu_{k})  &=~  \frac{1}{2} ||\nu_{k} ||^{2}_{2}  ~0 \leq k \leq N \nonumber  
\end {aligned}
\end {equation}

These specific loss function terms represent common choices in RNNs.
\\

\section {basic Recurrent Neural Networks (bRNN): summary equations}

We now summarize the set of equations to be used in coding computations as follows: 
\\

The basic RNN (bRNN) model is 
\begin {equation} 
\begin {aligned} 
x_{k+1} &= A x_{k} + U_{i}  h_{k} + W_{i}  s_{k} +  b_{i} , ~~ x_{o}, && ~ k=0,..., N-1 \\
 h_{k} &= \sigma_{k}  (x_{k} ), &&~ k=0,..., N  \\
 y_{k}  &= V_{i} h_{k} + D_{i}  s_{k} + c_{i} , &&~ k=0,..., N  
\end {aligned}
\end {equation}
where the adaptive parameters are indexed with the iteration (i.e., epoch) index $i$ to denote update of the parameters during training. These parameters are expected to converge to a constant at the end of the training process. The time index $k$ denotes the sample trajectory over the finite horizon training sequence, i.e., input-label pairs ($ s_k, ~ d_k$),  $0 \leq k \leq N$. 
We thus reserve the time index $k$ along each step of the sequence trajectory over the finite-horizon, $ o \leq k \leq N$, while the parameter update iteration index $i$ per full sequence trajectory as illustrated below: 
\\
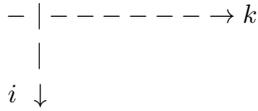
\begin {align}
      - &~|------\rightarrow k  \nonumber  \\
        &~|          \nonumber  \\
 i ~     &\downarrow           \nonumber  
\end {align}
\\
For each iteration $i$, this model propagates forward for a given (or chosen) initial condition $x_o$, and a sequence pair ($s_k$,  $d_k$) to generate the state $x_k$, the corresponding hidden nonlinearity $h_k$ and the output $y_k$ over the sequence duration $k=0, ..., N$. Thus the output error sequence is computed.  
The error backpropagation co-state equations need the final co-state and the output error sequence to generate the co-state dynamics backward in time $k$. Here, the final co-state is expressed as (see, \ref {eq: eqn4403}):
\begin {align}
\lambda_{N}  ~&=~ ( \sigma_{N}^{'})^T    V_{i}^{T} ~ (y_{N} - d_{N})   \nonumber  
\end {align}
\\
where $\sigma_{N}^{'}$ is the derivative of the nonlinear vector $\sigma_{N}$ which is represented as a diagonal matrix of element-wise derivatives of the corresponding scalar nonlinearity. It may also be represented as a vector of derivatives which point-wise multiplies to the error vector as a Hadamard vector-vector multiplication. Usually, in codes, the latter representation is followed. However, for analysis, it is more convenient to view it as a diagonal matrix (transpose) as it follows from the vector calculus. Let us compactly express the output error signal over the duration of $k$ as: 
\begin {align}
e_{k}  ~=~  (y_{k} - d_{k}) , ~~ 0 \leq k \leq N
\end {align}
Then the final time boundary condition of the co-state is expressed as: 
\begin {align}
\lambda_{N}  ~&=~ ( \sigma_{N}^{'})^T    V_{i}^{T} ~ e_{N}   \label {eq: eqn901}
\end {align}
\\
Now the back propagating co-state dynamics are (see Appendix II)
\begin {align}
\lambda_{k} =& ( A + U_{i}  \sigma_{k}^{'} )^T  \lambda_{k+1} ~+~
(\sigma_{k}^{'})^T    V_{i}^{T} ~ e_{k}  ~+  \nonumber  \\ 
&~\beta_1 ~ \left(\frac{\partial L^{k} }{\partial x_{k}}  \right)  ~+~   
\beta_2 ~(\sigma_{k}^{'})^T  \left(\frac{\partial L^{k} }{\partial h_{k}}  \right) ,   \nonumber \\  \label {eq: eqn902}
&~~~~~~~~~~~~~~~~~~~~~~~~~~~~~~~~~~~0 < k   \leq N-1
\end {align}

From Appendix II, the state equations parameter gradient changes at each time $k, 0 \leq k \leq N-1$ (without regularization) are: 
\begin {align}
\Delta U_{k} ~=~ - \eta  \lambda_{k+1} (h_{k})^{T}  \label {eq: eqn1101}
\end {align}
\begin {align}
\Delta W_{k} ~=~ - \eta   \lambda_{k+1} (s_{k})^{T}   \label {eq: eqn1102}
\end {align}
\begin {align}
\Delta b_{k} ~=~ - \eta    \lambda_{k+1}    \label {eq: eqn1103}
\end {align}
\\
Similarly, from Appendix II, the parameter (weight and bias) updates in the output equation at each time $k, 0 \leq k \leq N$ (without regularization) are: 
\begin {align}
\Delta V_{k} ~=~ - \eta  \left(\frac{\partial L^{k} }{\partial V_{k}} \right)  ~=~ - \eta  e_{k}  (h_{k})^{T}
 \label {eq: eqn1104}
\end {align}
\begin {align}
\Delta D_{k} ~=~ - \eta  \left(\frac{\partial L^{k} }{\partial D_{k}} \right)  ~=~   - \eta  e_{k}  (s_{k})^{T}
 \label {eq: eqn1105}
\end {align}
\begin {align}
\Delta c_{k} ~=~ - \eta  \left(\frac{\partial L^{k} }{\partial c_{k}} \right) ~=~    - \eta e_{k}  
 \label {eq: eqn1106}
\end {align}
\\
Usually, these parameter change contributions at each $k$ are summed over $k$ over their time horizon to provide one change per trajectory duration or epoch. Then, one may updates all the parameters at each single iteration (i.e. epoch) or over a patch or several iterations. 

Thus, per sequene trajectory, one can accumulate the parameter change contributions to obtain their updates as: 
 
\begin {align}
\Delta U_{i} ~=~ - \eta  \sum_{k=o}^{N-1} \lambda_{k+1} (h_{k})^{T}  
 \label {eq: eqn2101}
\end {align}

\begin {align}
\Delta W_{i} ~=~ - \eta   \sum_{k=o}^{N-1} \lambda_{k+1} (s_{k})^{T}  
 \label {eq: eqn2102}
\end {align}

\begin {align}
\Delta b_{i} ~=~ - \eta   \sum_{k=o}^{N-1}  \lambda_{k+1}   
 \label {eq: eqn2103}
\end {align}
\\
Similarly, the parameter (weight and bias) updates in the output equation at one iteration (epoch) are: 

\begin {align}
\Delta V_{i} ~=~ - \eta   \sum_{k=o}^{N} \left(\frac{\partial L^{k} }{\partial V_{k}} \right)  ~=~ - \eta   \sum_{k=o}^{N} e_{k}  (h_{k})^{T}
 \label {eq: eqn2104}
\end {align}

\begin {align}
\Delta D_{i} ~=~ - \eta  \sum_{k=o}^{N}  \left(\frac{\partial L^{k} }{\partial D_{k}} \right)  ~=~   - \eta   \sum_{k=o}^{N} e_{k}  (s_{k})^{T}
 \label {eq: eqn2105}
\end {align}

\begin {align}
\Delta c_{i} ~=~ - \eta   \sum_{k=o}^{N} \left(\frac{\partial L^{k} }{\partial c_{k}} \right) ~=~    - \eta  \sum_{k=o}^{N} e_{k}  
 \label {eq: eqn2106}
\end {align}
\\
\\
\textbf{\underline{Remark V.1:}}
The above parameter update expressions are commonly used forms, which are basically a sum of the contributions over the time index $k$ over the full sequence trajectory. We note that this is only one choice. Observe also that it is equivalent to using the \textit{mean} of the gradient changes over the sequences, 
eqns (\ref {eq: eqn1101})-(\ref{eq: eqn1106}), 
where $\eta$ absorbs any scaling due to the number of elements of the sequence. The mean, however, may be small or even zero while some gradient changes at some time indices $k$ may be relatively large. In fact, this is one explanation of the ``vanishing gradient" phenomena. That is, even though the co-state  and output error sequences are not vanishing, the gradient changes using the "mean" as in eqns (\ref {eq: eqn2101})-(\ref{eq: eqn2106}) are.
\\
\\
\textbf{\underline{Remark V.2:}}
Another choice to explore is to use the \textit{median} of the changes, or the \textit{minimum} of the changes, over the gradient change sequences, eqns (\ref {eq: eqn1101})-(\ref{eq: eqn1106}). Moreover, the change in power or strength in the sequence of changes is best captured by the variance which should influence the learning rate in a multitude of ways. Thus there are alternate possibilities of the iteration update laws for RNN besides the ones in eqns (\ref {eq: eqn2101})-(\ref{eq: eqn2106}). 
\\
\\
\textbf{\underline{Remark V.3:}}
The ``vanishing and exploding" gradients can be explained easily by investigating the update eqns 
(\ref {eq: eqn2101})-(\ref{eq: eqn2106}), and the backward co-state dynamics eqns 
(\ref {eq: eqn901})-(\ref{eq: eqn902}). By their nature, the state and co-state equations form an unstable saddle point dynamics, i.e. if the forward network dynamics are (locally) stable, the backward sensitivity dynamics are also stable but only backward in time.  This is another justification for the error back-propagation in addition to the fact that we use the  final co-state  value. In contrast, if the forward network dynamics are (locally) unstable, the backward sensitivity dynamics are also unstable backward in time.  In that event, the state will grow as it propagates forward, thus, the final costate will be large, and in turn the co-state will be further growing. As the co-state and the error signals are used in the expression of the gradient updates. the gradient will continue to grow unbounded. This process will make the state and co-state grow unbounded. As the time horizon increases (i.e., $N$ increases),  this will amplify the growth of the state and co-state leading to the so-called ``exploding gradient" phonemena. 

In the update laws and the iterations, one must use sufficiently small $\eta$ to avoid the occurence of numerical instability while allow the iteration process of the parameter update to be sufficiently small. The presence of the dominant stable linear term in bRNN ensures that the network for each bounded parameters is BIBO stable and thus the states, costates, and consequently the gradient would not grow unbounded. The bRRN can steer the error and the  co-state sequences towards zero during training. This is  possible of course for all co-states except for the co-state $\lambda_0$ which can be non-zero and does not play any role in the (state equations) parameter updates, eqn (\ref {eq: eqn2101})-(\ref{eq: eqn2103}).

\section{Conclusion}
This work introduces a design for a basic recurrent neural network (bRNN) which has a basic recurrent structure and output layer with sufficient parameters to enable a flexible filtering suitable for RNN applications of sequence to sequence mappings. The framework adopts the classical constrained optimization and calculus of variations to derive the backpropagation and (stochastic) gradient parameter update. It enables an ease in incorporating general loss function components on any variable in the networks, namely outputs, states or hidden units (nonlinear functions of the states). It shows the correspondence of the backpropagations through time (BTT) approach with applying the classical Lagrange multiplier method in constrained optimization. It further shows that the usual sum of contributions of changes to the parameters inherent in the BTT approach is only one form of update, and it could be a source of the ``vanishing gradient" phenomena. The ``exploding gradient" phenomena is explained to emanate from either(i) numerical instability due to summing contributions over a long time horizon, or (ii) due to the instability of the networks and the costate dynamics manivesting in the gradient update laws. Finaly, we state that we have conducted simulations using this bRRN model to verify and demostrate its perfromance and will be reported on the details of these results in another publication. \cite{IEEEhowto:ZZhou_2016}.


\section*{Acknowledgment}

The work is supported in part by the National Science foundation grant No. ECCS-1549517. The author thankfully acknowledges this support.
\\

\section{APPENDIX I: BIBO stability of bRNN}

We show that the presence of the linear term of the constant stable matrix in the basic recurrent neural network (bRNN) model as in eqn(\ref  {eq: eqn11}) is crucial for the boundedness of all trajectories, and that all trajectories would remain within, or converge to, a bounded region. This guarantees that, for bounded input signals and bounded parameters, trajectories would not go to infinity but rather remain confined to, or converge to,  a bounded region. 
\\
\\
The bRNN stability and dynamic behavior are governed by the dynamic eqn(\ref  {eq: eqn11}). To pursue the Bounded-Input-Bounded-Output (BIBO) stability analysis, we define the compact assumedly bounded vector: 
\begin {align} 
M_k~:=~ U  h_{k} + W  s_{k} +  b 
\end {align}
We choose the constant matrix A to be a stable matrix with eigenvalues having moduli less than unity.

Then, eqn(\ref  {eq: eqn11}) is re-written as
\begin {align} 
x_{k+1} &= A x_{k} + M_k , &&~~ k=0,..., N-1 ~ \label {eq: eqna11} 
\end {align}

For stability, we choose the following general quadratic form as a candidate Liapunov function: 
\begin {align}
V_{k} ~=~ (x_k+x^*_1)^T~ S ~(x_k+x^*_1)
\end {align}
\\
where $S$ is a symmetric positive definite matrix, and $x^*_1$ is a center point which may be different from the origin. In Liapunov theory, typically, $x^*_1$ is chosen to be an equilibrium point of interest. In the present case, however,  $x^*_1$  is simply a point in (the state) space as a center to be determined. We note that this Liapunov candidate is positive definite for all points with reference to the center point. 
\\
\\
Now, we calculate the ``difference equation" of this Liapunov function, i.e., 
\begin {align}
\Delta V_{k} &=~ V_{k+1} - V_k      \nonumber \\
&=~(x_{k+1}+x^*_1)^T S  (x_{k+1} + x^*_1) -  ~\label {eq: eqnaIII11} \nonumber \\ 
&~~~~~(x_k +x^*_1)^T S  (x_k+ x^*_1)
\end {align}
We use the dynamic equations, eqn(\ref {eq: eqna11}) into the equality (\ref {eq: eqnaIII11}), and expand terms to obtain a quadratic equation in $x_k$. We then complete the square and match the expanded equation terms to a quadratic expression term-by-term. We thus obtain the following general quadratic form: 
\begin {align}
\Delta V_{k} &=~- 
\left[G x_k -  x^*_2 \right]^T  \left[G x_k -  x^*_2 \right]  + ~D \nonumber \\  
\label {eq: eqnaI11} 
\end {align}
where the matrix $G$ satisfies the equality
\begin {align}
G^T G = S- A^T S A \label {eq: eqnaIIII11}
\end {align}
Thus the difference Liapunov function will be negative definite outside the ellipsoid defined by 
eqn (\ref {eq: eqnaI11}) with the left-hand side set to zero. This summarizes the general approach. 
\\
\\
To be specific, and to simplify the calculations, we choose $S=I$ (the identity matrix),  the vector $x^*_1= 0$, and proceed as follows. Eqn (\ref {eq: eqnaIIII11}) becomes
\begin {align}
G^T G = I- A^T A= Q^T [I- \Lambda^T \Lambda ] Q
\end {align}
Note that $G^T G$ is positive definite and thus invertible. 
$\Lambda $ is a diagonal matrix of the square roots of the singular values of $A$ with moduli less than 1, and $Q$ is the matrix formed of the corresponding orthogonal right-singular vectors. Note that the singular values of $A$ result in  
\begin {align}
A^T A= Q \Lambda^T   \Lambda Q^T
\end {align}

Then to match eqn (\ref{eq: eqnaI11}) to  eqn (\ref{eq: eqnaIII11}), 
term by term, the following equalities must be satisfied: 
\begin {align}
G^T x^*_2 ~&=~   A^T M_k  \\
(x^*_2)^T x^*_2~&=~-  M_k^TM_k + D
\end {align}
The above results in the following solution equalities
\begin {align}
x^*_2 ~&=~(G G^T)^{-1} G   A^T M_k  \\
D ~&=~ M_k^TM_k + (x^*_2)^T x^*_2
\end {align}
Thus all terms are well-defined in the difference eqn (\ref {eq: eqnaI11}). This completes the proof. 
Thus, we indeed found a Liapunov function candidate centered around $ x^*_1=0$  and is described by a spherical (or more generally, an ellipsoidal) region. Such region can be made to conservatively include the point  $ x^*_2$ where outside this bounded region the difference equation along  trajectories is negative definite.  
\\

We remark that our purpose here is to simply determine a bounded region where BIBO stability is ensured. In general, and for a tighter bound constraints, the goal is to solve the general constraints in order to determine a point $ x^*_1$ as near as possible to a point $ x^*_2$ where both satisfy the constraints. 
\\

Finally, we remark that, in the training process of RNNs, the parameters are frozen over the finite-horizon forward state dynamics and backward co-state dynamics. This constitutes one iteration (or epoch). The gradient descent update of the parameters occur per one or many iterations over the forzen training sequences. This constitutes a decoupling of the dynamics of the bRNN and the parameter gradient updates. Thus the BIBO stability of the overall system is maintained.

\section{APPENDIX II: Derivations}

We now provide specific details of the model and its parameter updates. We assume the state iniital condition $x_{o}$  is chosen or given as a constant, and thus $dx_{o}=0$.
The split boundary conditions then are: 
\begin {align}
x_{o}, ~~~
\lambda_{N} = \frac{\partial  \phi}{\partial x_{N}}   \label {eq: eqnaII11} 
\end {align}

Thus, the processing is to use the initial state $x_{o}$, the input signal sequence $s_{k}$ and the coressponding desired sequence $d_{k}$ to generate the  internal state and output sequences. This consequently generates the output error sequnece expressed compacted as: 
\begin {align}
e_k=  (y_{k} -d_{k}), ~ 0 \leq k \leq N
\end {align}
\\
The final-time loss function $ \phi (y_{N})$ is defined as an $L_{2}$ norm in the output error as:
\begin {align}
\phi (y_{N}) &= \frac{1}{2} ||y_{N} -d_{N}||^{2}_{2} \nonumber  \\
&= \frac{1}{2} e_{N}^T e_{N}
\end {align}
And thus its derivative in eqn (\ref {eq: eqnaII11}) provides the final co-state as 
\begin {align}
\lambda_{N} = \frac{\partial  \phi}{\partial x_{N}}  ~&=~  (\sigma_{N}^{'})^T    V_{N}^{T} ~ e_{N} 
\end {align}

The general loss function is simplified to be a sum of separate loss functions as 
\begin {align}
L^{k} (y_{k}, x_{k} , h_{k} ,  \theta_{k} , \nu_{k} ) ~&=~ L^{k} (y_{k})  \nonumber  \\
&~+ ~\beta L^{k} ( x_{k}) ~+~ \beta_{0} L^{k} ( h_{k}) \nonumber  \\
&~ +~\gamma_1  L^{k} (  \theta_{k}) ~+~ \gamma_2  L^{k} (  \nu_{k})
\end {align}

We now calculate the derivatives needed in the co-state equation (\ref {eq: eqn303}) and in the gradient descent parameter update equations (\ref {eq: eqn101}) and (\ref {eq: eqn102}),  specilized to each parameter. 
\\

\begin {align}
\frac{\partial  L^{k}}{\partial x_{k}}  &=~\frac{\partial  L^{k} (y_k) }{\partial x_{k}} +
\beta \frac{\partial  L^{k} (x_k) }{\partial x_{k}} +
\beta_0 \frac{\partial  L^{k} (h_k)}{\partial x_{k}}  \nonumber \\ 
&= (\sigma_{k}^{'})^T    V_k^{T} ~ e_{k} + 
\beta  \left(\frac{\partial L^{k} }{\partial x_{k}} \right) + \beta_0 (\sigma_{k}^{'})^T  \left(\frac{\partial L^{k} }{\partial h_{k}} \right),    \nonumber  \\ 
&~~~0 < k   \leq N-1
\end {align}
And the derivative (The Jacobian): 
\begin {align}
\left(\frac{\partial f^{k} }{\partial x_{k}} \right) ~=~ ( A + U_{k}  (\sigma_{k})^{'}) , ~~~0 < k   \leq N-1
\end {align}

Also, one derives the equalities: 

\begin {align}
\frac{\partial  L^{k} }{\partial \theta_{k}}  =  \gamma_1  \theta_{k}
\end {align}
\\
\begin {align}
\left(\frac{\partial f^{k} }{\partial U_{k}} \right)^{T}  \lambda_{k+1} ~=~   \lambda_{k+1} (h_{k})^{T} 
\end {align}

\begin {align}
\left(\frac{\partial f^{k} }{\partial W_{k}} \right)^{T}  \lambda_{k+1}  ~=~   \lambda_{k+1} (s_{k})^{T}
\end {align}

\begin {align}
\left(\frac{\partial f^{k} }{\partial b_{k}} \right)^{T}  \lambda_{k+1} ~=~   \lambda_{k+1}
\end {align}
\\

Thus, this gives: 
\begin {align}
\Delta U_{k} ~=~ - \eta   \left(   \gamma_1  U_{k}    +    \lambda_{k+1} (h_{k})^{T}   \right)
\end {align}

\begin {align}
\Delta W_{k} ~=~ - \eta   \left(   \gamma_1  W_{k}    +    \lambda_{k+1} (s_{k})^{T}   \right)
\end {align}

\begin {align}
\Delta b_{k} ~=~ - \eta   \left(   \gamma_1  b_{k}    +    \lambda_{k+1}    \right)
\end {align}
The parameter (weight and bias) updates in the output equation are now calculated. 
The regularization terms for all parameters contribute
\begin {align}
\frac{\partial  L^{k} (\nu)  }{\partial \nu_{k}}  =  \gamma_2  \nu_{k}
\end {align}
\\

Finally, the gradient changes at each time step $k$ are
\begin {equation}
\begin {aligned}
\Delta \nu_{k} = - \eta
\frac{\partial  H^{k}}{\partial \nu_{k}}   = - \eta
\left( \frac{\partial  L^{k}}{\partial \nu_{k}}   \right)
\end {aligned}
\end {equation}
where 

\begin {align}
\Delta V_{k} = - \eta  \left(\frac{\partial L^{k} }{\partial V_{k}} \right)  = - \eta 
\left(  \gamma_2  V_k +   e_{k}  (h_{k})^{T}  \right)
\end {align}

\begin {align}
\Delta D_{k} = - \eta  \left(\frac{\partial L^{k} }{\partial D_{k}} \right)  =   - \eta  
\left( \gamma_2  D_k +  e_{k}  (s_{k})^{T}  \right)
\end {align}

\begin {align}
\Delta c_{k} = - \eta  \left(\frac{\partial L^{k} }{\partial c_{k}} \right) =    - \eta
 \left( \gamma_2  c_k +  e_{k}   \right)
\end {align}
\\

Thus, the (training) processing is to use the initial state $x_{o}$, and the input signal sequence $s_{k}$ and its  corresponding desired sequence $d_{k}$ to generate the  gradient descent parameter changes. Then one updates the parameter online, over the full sequence (epoch), or over several epoches, to steer the parameters along a stochastic gradient descent towards a good local optimal. 
 



\bibliographystyle{IEEEtran}

\begin{thebibliography}{1}

\bibitem{IEEEhowto:lecun_deep_2015}
{Y. {LeCun}, Y.  Bengio, and G. Hinton},  \emph{Deep learning}, 
\hskip 1em plus 0.5em minus 0.4em\relax Nature, vol. {521}, pp. 436-444, May 2015.
\bibitem{IEEEhowto:Chung_2014}
{Chung, Junyoung and Gulcehre, Caglar and Cho, {KyungHyun} and Bengio, Yoshua},  \emph{Empirical Evaluation of Gated Recurrent Neural Networks on Sequence Modeling},
 \hskip 1em plus 0.5em minus 0.4em\relax  {https://arxiv.org/abs/1412.3555}, 2014.
\bibitem{IEEEhowto:Joze_2015}  
R. Jozefowicz, W. Zaremba, and I. Sutskever,  \emph{An empirical exploration of recurrent network architectures}, 
\hskip 1em plus  0.5em minus 0.4em\relax 2015. 
\bibitem{IEEEhowto:Zhou_minimal_2016}
{G-B. Zhou,  J. Wu, Jianxin,  C-L. Zhang, and Z-H. Zhou}, \emph{Minimal Gated Unit for Recurrent Neural Networks},
\hskip 1em plus  0.5em minus 0.4em\relax  ArXiv: {https://arxiv.org/abs/1603.09420},  2016.
\bibitem{IEEEhowto:Le_2015}
Q. V. Le, N.  Jaitly,  and G. E. ~Hinton,  \emph{A simple way to initialize recurrent networks of rectified linear units}, \hskip 1em plus 0.5em minus 0.4em\relax {{arXiv} preprint {arXiv}:1504.00941}, [cs.NE] 7 April 2015.
\bibitem{IEEEhowto:Bryson_1975}
A. E. Bryson, Jr and Y-C. Ho,  \emph{Applied Optimal Control: Optimization, Estimation and Control},  \hskip 1em plus
0.5em minus 0.4em\relax  CRC Press, 1975
\bibitem{IEEEhowto:Lewis_2012}  
F. Lewis, D. Vrabie, and V. Syrmos, \emph{Optimal Control}, 3rd ~ed \hskip 1em plus
0.5em minus 0.4em\relax Wiley 2012
\bibitem{IEEEhowto:Khuram_2003}  
K. Waheed, F. M. Salem,  \emph{Blind source recovery: A framework in the state space}, 
\hskip 1em plus 0.5em minus 0.4em\relax 
Journal of Machine Learning Research 4 (Dec), pp. 1411-1446.
\bibitem{IEEEhowto:Albatain_2016} 
Z. Albataineh and F. M. Salem, 
\emph{Adaptive Blind CDMA Receivers Based on ICA Filtered Structures},
\hskip 1em plus 0.5em minus 0.4em\relax 
 Circuits Syst Signal Process (2016). doi:10.1007/s00034-016-0459-4.
\bibitem{IEEEhowto:ZZhou_2016} 
Z. Zhou and F. M. Salem, 
\emph{Perfomance Evaluation of the basic RNN model},
\hskip 1em plus 0.5em minus 0.4em\relax 
Memorandum.10.12.2016. MSU, Dec. 2016
\end{thebibliography}
%

\end{document}